\documentclass[11pt,a4paper]{article}
\usepackage[hyperref]{acl2020}
\usepackage{times}
\usepackage{latexsym}

\usepackage{mystyle}
\usepackage{symbol}
\usepackage{amsmath}
\usepackage{lipsum}
\usepackage{mathtools}
\usepackage{cuted}
\usepackage{microtype}

\aclfinalcopy 
\def\aclpaperid{2042} 

\newcommand{\theTitle}{Structured Tuning for Semantic Role Labeling}

\title{\theTitle}
\author{Tao Li \\
  University of Utah \\
  \texttt{tli@cs.utah.edu} \\\And
  Parth Anand Jawale \\
  University of Colorado \\
  \texttt{Parth.Jawale@colorado.edu} \AND
  Martha Palmer \\
  University of Colorado \\
  \texttt{Martha.Palmer@colorado.edu} \\\And
  Vivek Srikumar \\
  University of Utah \\
  \texttt{svivek@cs.utah.edu}}

\date{}

\newcommand{\lbl}[1]{\texttt{#1}}

\newcommand{\skippedDetails}[1]{}

\begin{document}
\maketitle

\begin{abstract}
  Recent neural network-driven semantic role labeling (SRL) systems have shown impressive improvements in F1 scores. These improvements are due to expressive input representations, which, at least at the surface, are orthogonal to knowledge-rich constrained decoding mechanisms that helped linear SRL models. Introducing the benefits of structure to inform neural models presents a methodological challenge.  In this paper, we present a structured tuning framework to improve models using softened constraints only at training time.  Our framework leverages the expressiveness of neural networks and provides supervision with structured loss components.  We start with a strong baseline (RoBERTa) to validate the impact of our approach, and show that our framework outperforms the baseline by learning to comply with declarative constraints.  Additionally, our experiments with smaller training sizes show that we can achieve consistent improvements under low-resource scenarios.
\end{abstract}


\section{Introduction}
\label{sec:intro}

Semantic Role Labeling~\cite[SRL,][]{palmer2010semantic} is the task of labeling
semantic arguments of predicates in sentences to identify who does what to
whom. Such representations can come in handy in tasks involving text
understanding, such as coreference
resolution~\cite{ponzetto-strube-2006-exploiting} and reading
comprehension~\cite[e.g.,][]{berant-etal-2014-modeling,zhang2019semantics}. This
paper focuses on the question of how knowledge can influence modern semantic
role labeling models.

Linguistic knowledge can help SRL models in several ways. For example, syntax
can drive feature design~\cite[e.g.,][ and
others]{punyakanok2005necessity,toutanova-etal-2005-joint,kshirsagar-etal-2015-frame,johansson-nugues-2008-dependency},
and can also be embedded into neural network
architectures~\cite{strubell2018linguistically}.

In addition to such influences on input representations, knowledge about the nature of semantic roles can inform structured decoding algorithms used to construct the outputs. The SRL literature is witness to a rich array of techniques for structured inference, including integer linear programs~\cite[e.g.,][]{punyakanok2005necessity,punyakanok2008importance}, bespoke inference algorithms~\cite[e.g.,][]{tackstrom2015efficient}, A* decoding~\cite[e.g.,][]{he-etal-2017-deep}, greedy heuristics~\cite[e.g.,][]{ouchi-etal-2018-span}, or simple Viterbi decoding to ensure that token tags are BIO-consistent.

By virtue of being constrained by the definition of the task, global inference promises semantically meaningful outputs, and could provide valuable signal when models are being trained. However, beyond Viterbi decoding, it may impose prohibitive computational costs, thus ruling out using inference during training. Indeed, optimal inference may be intractable, and inference-driven training may require ignoring certain constraints that render inference difficult. 


While global inference was a mainstay of SRL models until recently, today's end-to-end trained neural architectures have shown remarkable successes without needing decoding.  These successes can be attributed to the expressive input and internal representations learned by neural networks.  The only structured component used with such models, if at all, involves sequential dependencies between labels that admit efficient decoding.

In this paper, we ask: \emph{Can we train neural network models for semantic roles in the presence of general output constraints, without paying the high computational cost of inference?}
We propose a structured tuning approach that exposes a neural SRL model to
differentiable constraints during the finetuning step.
To do so, we first write the output space constraints as logic rules.  Next, we
relax such statements into differentiable forms that serve as
regularizers to inform the model at training time.  Finally, during inference,
our structure-tuned models are free to make their own judgments about labels
without any inference algorithms beyond a simple linear sequence decoder.

We evaluate our structured tuning on the CoNLL-05~\cite{carreras-marquez-2005-introduction} and CoNLL-12 English SRL~\cite{pradhan-etal-2013-towards}
shared task datasets, and show that by learning to comply with declarative constraints,
trained models can make more consistent and more accurate predictions.  We
instantiate our framework on top of a strong baseline system based on the
RoBERTa~\cite{liu2019roberta} encoder, which by itself performs on par with previous best SRL models that are not ensembled.
We evaluate the impact of three different types of
constraints.  Our experiments on the CoNLL-05 data show that our constrained
models outperform the baseline system by $0.2$ F1 on the WSJ section and $1.2$
F1 on the Brown test set.  Even with the larger and cleaner CoNLL-12 data, our
constrained models show improvements without introducing any additional
trainable parameters.  Finally, we also evaluate the effectiveness of our
approach on low training data scenarios, and show that constraints can be more
impactful when we do not have large training sets.

In summary, our contributions are:
\begin{enumerate}[nosep]
\item We present a structured tuning framework for SRL which uses soft
  constraints to improve models without introducing additional trainable parameters.\footnote{Our code to replay our
      experiments is archived at \url{https://github.com/utahnlp/structured_tuning_srl}.}
\item Our framework outperforms strong baseline systems, and shows especially large improvements
  in low data regimes.
\end{enumerate}


\section{Model \& Constraints}

In this section, we will introduce our structured tuning framework for semantic
role labeling.  In \S\ref{sec:baseline}, we will briefly cover the baseline system.  To that, we will add three constraints, all treated as combinatorial constraints requiring inference algorithms in past work: \textbf{Unique Core Roles} in \S\ref{sec:uniqueconstr}, \textbf{Exclusively Overlapping Roles} in \S\ref{sec:overlapconstr}, and \textbf{Frame Core Roles} in \S\ref{sec:frameconstr}.  For each constraint, we will discuss how to use its softened version during training.

We should point out that the specific constraints chosen serve as a
proof-of-concept for the general methodology of tuning with declarative
knowledge. For simplicity, for all our experiments, we use the ground truth
predicates and their senses. 

\subsection{Baseline}\label{sec:baseline}
We use RoBERTa~\cite{liu2019roberta} base version to develop our baseline SRL system. The large number of parameters not only allows it to make fast and accurate predictions, but also offers the capacity to learn from the rich output structure, including the constraints from the subsequent sections.

Our base system is a standard BIO tagger, briefly outlined below. Given a sentence $s$, the goal is to assign a label of the form \lbl{B-X}, \lbl{I-X} or \lbl{O} for each word $i$ being an argument with label \lbl{X} for a predicate at word $u$. These unary decisions are scored as follows:
\begin{align}
	e &= \textsl{map}(\text{RoBERTa}(s))\\
	v_u, a_i &= f_v(e_u), f_a(e_i)\\
	\phi_{u,i} &= f_{va}([v_u, a_i])\\
	y_{u,i} &= g(\phi_{u,i})
\end{align}
Here, 
$\textsl{map}$ converts the wordpiece embeddings $e$ to whole word embeddings by summation,
$f_v$ and $f_a$ are linear transformations of the predicate and argument embeddings respectively,
$f_{va}$ is a two-layer ReLU with concatenated inputs,
and finally $g$ is a linear layer followed by softmax activation that predicts a probability distribution over labels for each word $i$ when $u$ is a predicate.
In addition, we also have a standard first-order sequence model over label sequences for each predicate in the form of a CRF layer that is Viterbi decoded.
We use the standard cross-entropy loss to train the model. 

\subsection{Designing Constraints}\label{sec:designconstr}
Before looking at the specifics of individual constraints, let us first look at a broad overview of our methodology. We will see concrete examples in the subsequent sections.

Output space constraints serve as prior domain knowledge for the SRL task.
We will design our constraints as invariants at the training stage.
To do so, we will first define constraints as statements in logic.
Then we will systematically relax these Boolean statements into differentiable forms using concepts borrowed from the study of triangular norms ~\cite[t-norms,][]{klement2013triangular}.
Finally, we will treat these relaxations as regularizers in addition to the standard cross-entropy loss.

All the constraints we consider are conditional statements of the form:
\begin{align}
	\forall x,  L(x) \rightarrow R(x) \label{eq:toprule}
\end{align}
where the left- and the right-hand sides---$L(x), R(x)$ respectively---can be either disjunctive or conjunctive expressions.
The literals that constitute these expressions are associated with classification neurons, \ie, the predicted output probabilities are soft versions of these literals.

What we want is that model predictions satisfy our constraints.
To teach a model to do so, we transform conditional statements into regularizers,
such that during training, the model receives a penalty if the rule is not satisfied for an example.\footnote{Constraint-derived regularizers are dependent on examples, but not necessarily labeled ones. For simplicity, in this paper, we work with sentences from the labeled corpus. However, the methodology described here can be extended to use unlabeled examples as well.}

To soften logic, we use the conversions shown in Table~\ref{tab:translation} that combine the product and G\"odel t-norms.
We use this combination because it offers cleaner derivatives make learning easier.
A similar combination of t-norms was also used in prior work~\cite{minervini2018adversarially}.
Finally, we will transform the derived losses into log space to be consistent with cross-entropy loss. \citet{li2019consistency} outlines this relationship between the cross-entropy loss and constraint-derived regularizers in more detail.

\begin{table}[ht!]
  \centering
  \setlength{\tabcolsep}{3.5pt}
  \begin{tabular}{lcccc}
    \toprule
    Logic & $\bigwedge_i a_i$ & $\bigvee_i a_i$ & $\neg a$ & $a\rightarrow b$\\
    \midrule  
    G\"odel & $\min\p{a_i}$ & $\max\p{a_i}$ & $1-a$ & -- \\
    Product & $\Pi a_i$ & -- & $1-a$ & $\min\p{1, \frac{b}{a}}$\\
    \bottomrule
  \end{tabular}
  \caption{Converting logical operations to differentiable forms.
  For literals inside of $L(s)$ and $R(s)$, we use the G\"odel t-norm.
  For the top-level conditional statement, we use the product t-norm.
  Operations not used this paper are marked as `--'.}
  \label{tab:translation}
\end{table}

\subsection{Unique Core Roles ($U$)}\label{sec:uniqueconstr}

Our first constraint captures the idea that, in a frame, there can be at most one core participant of a given type. Operationally, this means that for every predicate in an input sentence $s$, there can be no more than one occurrence of each core argument (i.e, $\mathcal{A}_{core} = \{\lbl{A0}, \lbl{A1}, \lbl{A2}, \lbl{A3}, \lbl{A4}, \lbl{A5}\}$). In first-order logic, we have:
\begin{align}
	&\forall~~u,i \in s, \lbl{X} \in \mathcal{A}_{core}, \nonumber\\
	&\quad\quad B_{\lbl{X}}(u, i) \rightarrow \bigwedge_{j\in s,j\neq i} \neg B_{\lbl{X}}(u, j) \label{eq:uniqueconstr}
\end{align}
which says, for a predicate $u$, if a model tags the $i$-th word as the beginning of the core argument span,
then it should not predict that any other token is the beginning of the same label.

In the above rule, the literal $B_{\lbl{X}}$ is associated with the predicted probability for the label \lbl{B-X}\footnote{
We will use $B_{\lbl{X}}(u, i)$ to represent both the literal that the token $i$ is labeled with \lbl{B-X} for predicate $u$ and also the probability for this event. We follow a similar convention for the \lbl{I-X} labels.}.
This association is the cornerstone for deriving constraint-driven regularizers. Using the conversion in Table~\ref{tab:translation} and taking the natural $\log$ of the resulting expression, we can convert the implication in \eqref{eq:uniqueconstr} as $l(u, i, \lbl{X})$:

{\small
\begin{align*}
  \max\left( \log B_{\lbl{X}}\p{u, i} - \min_{j\in s,j\neq i} \log\p{1-B_{\lbl{X}}\p{u, j}}\right).
\end{align*}
}
Adding up the terms for all tokens and labels, we get the final regularizer $L_{U}(s)$:
\begin{align}
  L_{U}(s) = \sum_{(u,i)\in s,\lbl{X}\in \mathcal{A}_{core}} l(u, i,\lbl{X}).\label{eq:uniqueloss}
\end{align}
Our constraint is universally applied to all words and predicates (\ie, $i, u$ respectively) in the given sentence $s$.
Whenever there is a pair of predicted labels for tokens $i, j$ that violate the rule~\eqref{eq:uniqueconstr}, our loss will yield a positive penalty.

\paragraph{Error Measurement $\rho_u$} To measure the violation rate of this constraint,
we will report the percentages of propositions that have duplicate core arguments.
We will refer to this error rate as $\rho_u$.

\subsection{Exclusively Overlapping Roles ($O$)}\label{sec:overlapconstr}
We adopt this constraint from~\citet{punyakanok2008importance} and related work.
In any sentence, an argument for one predicate can either be contained in or entirely outside another argument for any other predicate.
We illustrate the intuition of this constraint in Table~\ref{tab:nonoverlap}, assuming core argument spans are unique and tags are BIO-consistent.

\begin{table}[h]
  \centering
  \begin{tabular}{lcccc}
    \toprule
    \small Token index & $i$                                        & $\cdots$ & $j$           & $j+1$              \\\midrule  
    \small{$[i$-$j]$ has label \lbl{X}}  & $B_{\lbl{X}}$                              & $\cdots$ & $I_{\lbl{X}}$ & $\neg I_{\lbl{X}}$ \\
    \small{Not allowed} & --                                         & --       & $B_{\lbl{Y}}$ & $I_{\lbl{Y}}$      \\
    \small{Not allowed} & $\neg B_{\lbl{Y}} \wedge \neg I_{\lbl{Y}}$ & --       & $I_{\lbl{Y}}$ & $I_{\lbl{Y}}$      \\
    \bottomrule
  \end{tabular}
  \caption{Formalizing the exclusively overlapping role constraint in terms of
    the $B$ and $I$ literals.  For every possible span $[i$-$j]$ in a sentence,
    whenever it has a label \lbl{X} for some predicate (first row), token labels
    as in the subsequent rows are not allowed for any other predicate for any
    other argument \lbl{Y}. Note that this constraint does not affect the cells
    marked with a --.}
  \label{tab:nonoverlap}
\end{table}

Based on Table~\ref{tab:nonoverlap}, we design a constraint that says:
if an argument has boundary $[i,j]$, then no other argument span can cross the boundary at $j$. This constraint applies to all argument labels in the task, denoted by the set $\mathcal{A}$.
\begin{align}
	\forall~ & u,i,j \in s \text{ such that } j > i, \text{and } \forall~\lbl{X}\in\mathcal{A}, \nonumber                                                                        \\
                 & P(u, i, j, \lbl{X}) \rightarrow \mathop{\bigwedge_{v\in s, \lbl{Y}\in\mathcal{A}}}_{(u,\lbl{X}) \neq (v,\lbl{Y})} Q(v, i, j, \lbl{Y}) \label{eq:nonoverlapconstr} \\
                 & \text{where } \nonumber                                                                                                                                           \\
	\small\begin{split}
                 & P(u, i, j, \lbl{X})= B_{\lbl{X}}(u, i) \wedge I_{\lbl{X}}(u, j) \wedge \neg I_{\lbl{X}}(u, j+1) \nonumber                                                         \\
                 & Q(v, i, j, \lbl{Y})  = Q_1(v, i, j, \lbl{Y}) \wedge Q_2(v, i, j, \eta)\nonumber                                                                                   \\
                 & Q_1(v, i, j, \lbl{Y})= \neg B_\lbl{Y}(v, j) \vee \neg I_\lbl{Y}(v, j+1)  \nonumber                                                                                \\
                 & Q_2(v, i, j, \lbl{Y})= \nonumber                                                                                                                                  \\
                 & \hspace{1em} B_\lbl{Y}(v, i)\vee I_\lbl{Y}(v, i) \vee \neg I_\lbl{Y}(v,j) \vee \neg I_\lbl{Y}(v, j+1)  \nonumber
	\end{split}
\end{align}
Here, the term $P(u, i, j, \lbl{X})$ denotes the indicator for the argument span $[i,j]$ having the label $\lbl{X}$ for a predicate $u$ and corresponds to the first row of Table~\ref{tab:nonoverlap}.
The terms $Q_1(v, i, j, \lbl{Y})$ and $Q_2(v, i, j, \lbl{Y})$ each correspond to prohibitions of the type described in the second and third rows respectively.

As before, the literals $B_{\lbl{X}}$, etc are relaxed as model probabilities to define the loss.
By combining the
G\"odel and product t-norms, we translate Rule~\eqref{eq:nonoverlapconstr} into:
\begin{align}
  L_{O}(s) = \mathop{\sum_{(u,i,j)\in s}}_{j > i, \lbl{X}\in\mathcal{A}} l(u,i,j,\lbl{X}).
\end{align}
\noindent where,
{\small
\begin{align*}
 & l(u,i,j, \lbl{X}) = \max\big(0,  \log P(u,i,j,\lbl{X})                                                                                         \\
 & \quad\quad\quad\quad\quad\quad\quad\quad\quad\quad- \mathop{\min_{v\in s, \lbl{Y}\in\mathcal{A}}}_{(u,\lbl{X}) \neq (v,\lbl{Y})} \log Q(v, i, j, \lbl{Y})\big) \\
 & P(u,i,j,\lbl{X}) =                                                                                                                                \\
 & \quad\min\p{B_{\lbl{X}}\p{u, i}, I_{\lbl{X}}\p{u, j}, 1-I_{\lbl{X}}\p{u, j+1}}                                                                    \\    
 & Q(v, i, j, \lbl{Y}) = \min\p{Q_1(v, i, j, \lbl{Y}), Q_2(v, i, j, \lbl{Y})}                                                                        \\
 & Q_1(v,i,j,\lbl{Y}) = 1 - \min\p{B_{\lbl{Y}}(v,j), I_{\lbl{Y}}(v, j+1)}                                                                            \\
 & Q_2(v,i,j,\lbl{Y}) =                                                                                                                              \\
 & \quad\max\p{B_{\lbl{Y}}(v, i), I_{\lbl{Y}}(v, i), 1-I_{\lbl{Y}}(v, j), 1-I_{\lbl{Y}}(v, j+1)} 
\end{align*}
}
Again, our constraint applies to all predicted probabilities.
However, doing so requires scanning over $6$ axes defined by $(u,v, i, j, \lbl{X}, \lbl{Y})$, which is computationally expensive.
To get around this, we observe that, since we have a conditional statement, the higher the probability of $P(u,i,j,\lbl{X})$, the more likely it yields non-zero penalty.
These cases are precisely the ones we hope the constraint helps.
Thus, for faster training and ease of implementation, we modify Equation~\ref{eq:nonoverlapconstr} by squeezing the $(i,j)$ dimensions using $\text{top-k}$ to redefine $L_{O}$ above as:
\begin{align}
  &\mathcal{T}(u, \lbl{X}) = \arg \text{top-k}_{\p{i,j}\in s} P\p{u, i,j, \lbl{X}} \\
  &L_{O}(s)= \sum_{u\in s,\lbl{X}\in\mathcal{A}} \sum_{(i,j)\in \mathcal{T}(v, \lbl{X})} l(u,i,j,\lbl{X}).
\end{align}
where $\mathcal{T}$ denotes the set of the top-k span boundaries for predicate $u$ and argument label $\lbl{X}$.
This change results in a constraint defined by $u$, $v$, \lbl{X}, \lbl{Y} and the $k$ elements of $\mathcal{T}$.

\paragraph{Error Measurement $\rho_o$} We will refer to the error of the overlap constraint as $\rho_o$,
which describes the total number of non-exclusively overlapped pairs of arguments.
In practice, we found that models rarely make such observed mistakes.
In \S\ref{sec:experiments}, we will see that using this constraint during training helps models
generalize better with other constraints.
In \S\ref{sec:topk}, we will analyze the impact of the parameter $k$ in the optimization described above.

\subsection{Frame Core Roles ($F$)}\label{sec:frameconstr}
The task of semantic role labeling is defined using the PropBank frame definitions. That is, for any predicate lemma of a given sense, PropBank defines which core arguments it can take and what they mean.
The definitions allow for natural constraints that can teach models to avoid predicting core arguments outside of the predefined set.
\begin{align}
	& \forall u\in s, k \in \mathcal{S}(u), \nonumber \\
	& \hspace{2ex} \text{Sense}(u, k) \rightarrow \mathop{\bigwedge_{i\in s}}_{\lbl{X} \not\in \mathcal{R}(u,k)} \neg \p{ B_{\lbl{X}}(u, i) \wedge I_{\lbl{X}}(u, i) } \nonumber
\end{align}
where $\mathcal{S}(u)$ denotes the set of senses for a predicate $u$, and 
$\mathcal{R}(u,k)$ denotes the set of acceptable core arguments when the predicate $u$ has sense $k$.

As noted in \S\ref{sec:designconstr}, literals in the above statement can to be associated with classification neurons.
Thus the $\text{Sense}(u,k)$ corresponds to either model prediction or ground truth.
Since our focus is to validate the  approach of using relaxed constraints for SRL, we will use the latter. 

This constraint can be also converted into regularizer following previous examples, giving us a loss term $L_{F}(s)$.

\paragraph{Error Measurement $\rho_f$} We will use $\rho_f$ to denote the violation rate.
It represents the percentage of propositions that have predicted core arguments outside the role sets of PropBank frames.

\paragraph{Loss} Our final loss is defined as:
\begin{align}
  &L_E(s) + \lambda_{U} L_{U}(s) + \lambda_{O} L_{O}(s) + \lambda_{F} L_{F}(s) \label{eq:finalloss}
\end{align}
Here, $L_E(s)$ is the standard cross entropy loss over the BIO labels, and the
$\lambda$'s are hyperparameters.


\section{Experiments \& Results}\label{sec:experiments}
In this section, we study the question: \emph{In what scenarios can we
  inform an end-to-end trained neural model with declarative knowledge?}  To
this end, we experiment with the CoNLL-05 and CoNLL-12 datasets, using standard
splits and the official evaluation script for measuring performance.  To empirically verify our framework in various data regimes, we
consider scenarios ranging from where only limited training data is available,
to ones where large amounts of clean data are available.

\subsection{Experiment Setup}
Our baseline (described in \S\ref{sec:baseline}) is based on RoBERTa.
We used the pre-trained base version released by~\citet{wolf2019transformers}.
Before the final linear layer, we added a dropout layer~\cite{srivastava2014dropout} with probability $0.5$. To capture the sequential  dependencies between labels, we added a standard CRF layer.
At testing time, Viterbi decoding with hard transition constraints was employed across all settings.
In all experiments, we used the gold predicate and gold frame senses.

Model training proceeded in two stages:
\begin{enumerate}[nosep]
\item We use the finetuned the pre-trained RoBERTa model on SRL with \emph{only} cross-entropy loss for $30$ epochs
with learning rate $3\times 10^{-5}$.
\item Then we continued finetuning with the combined loss in Equation~\ref{eq:finalloss} for another $5$ epochs with a lowered
learning rate of $1\times 10^{-5}$.
\end{enumerate}
During both stages, learning rates were warmed up linearly for the first $10\%$ updates. 

For fair comparison, we finetuned our baseline twice (as with the
constrained models); we found that it consistently outperformed the singly
finetuned baseline in terms of both error rates and role F1.  We grid-searched
the $\lambda$'s by incrementally adding regularizers.  The combination of
$\lambda$'s with good balance between F1 and error $\rho$'s on the dev set were
selected for testing.
We refer readers to the appendix for the values of $\lambda$'s.

For models trained on the CoNLL-05 data, we report performance on the dev set, and the WSJ and Brown test sets. For CoNLL-12 models, we report performance on the dev and the test splits.

\subsection{Scenario 1: Low Training Data}
Creating SRL datasets requires expert annotation, which is expensive. While there are some efforts on  semi-automatic annotation targeting low-resource languages~\citep[e.g.,][]{akbik-etal-2016-towards}, achieving high neural network performance with small or unlabeled datasets remains a challenge~\citep[e.g.,][]{furstenau-lapata-2009-graph,furstenau2012semi,titov-klementiev-2012-semi,gormley-etal-2014-low,abend-etal-2009-unsupervised}.


In this paper, we study the scenario where we have small amounts of fully labeled training data.
We sample $3\%$ of the training data and an equivalent amount of development examples.
The same training/dev subsets are used across all models.

Table~\ref{tab:resultlow}  reports the performances of using $3\%$ training data from
CoNLL-05 and CoNLL-12  (top and bottom respectively).
We compare our strong baseline model with structure-tuned models using all three
constraints. Note that for all these evaluations, while we use subsamples of the
dev set for model selection, the evaluations are reported using the full dev and
test sets.

We see that training with constraints greatly improves precision with low training data,
while recall reduces.
This trade-off is accompanied by a reduction in the violation rates $\rho_u$ and $\rho_f$.
As noted in \S\ref{sec:overlapconstr}, models rarely predict label sequences that violate the exclusively overlapping roles constraint.
As a result, the error rate $\rho_o$ (the number of violations) only slightly fluctuates.

\begin{table}[ht!]
  \centering
  \setlength{\tabcolsep}{3pt}
  \renewcommand{\arraystretch}{0.9}
  \begin{tabular}{lccc|c|ccc}
  	\multicolumn{7}{c}{\small{CoNLL-05 (3\%, 1.1k)}} \\
    \toprule
    \small{Dev} & \small{P} & \small{R} & \small{F1} & $\delta$\small{F1} & $\rho_u$ & $\rho_o$ & $\rho_f$ \\
    \midrule  
    \small{RoBERTa}$^2$ & \small{67.79} & \textbf{\small{72.69}} & \small{70.15} & & \small{14.56} & \small{23} & \small{6.19}  \\
    \small{+U,F,O}		& \textbf{\small{70.40}} & \small{71.91} & \textbf{\small{71.15}} & \small{1.0} & \small{8.56} & \small{20} &\small{5.82} \\
    \bottomrule
    \small{WSJ} & \small{P} & \small{R} & \small{F1} & $\delta$\small{F1} & $\rho_u$ & $\rho_o$ & $\rho_f$ \\
    \midrule  
    \small{RoBERTa}$^2$ & \small{70.48} & \textbf{\small{74.96}} & \small{72.65} & & \small{13.35} & \small{37} & \small{NA} \\
    \small{+U,F,O}		& \textbf{\small{72.60}} & \small{74.13} & \textbf{\small{73.36}} & \small{0.7} & \small{7.46} & \small{49} & \small{NA} \\
    \bottomrule
    \small{Brown} & \small{P} & \small{R} & \small{F1} & $\delta$\small{F1} & $\rho_u$ & $\rho_o$ & $\rho_f$ \\
    \midrule  
    \small{RoBERTa}$^2$ & \small{62.16} & \textbf{\small{66.93}} & \small{64.45} & & \small{12.94} & \small{6} & \small{NA} \\
    \small{+U,F,O}		& \textbf{\small{64.31}} & \small{65.64} & \textbf{\small{64.97}} & \small{0.5} & \small{5.47} & \small{6} & \small{NA} \\
    \bottomrule
    \bottomrule
    \multicolumn{7}{c}{\small{CoNLL-12 (3\%, 2.7k)}} \\
    \toprule
    \small{Dev} & \small{P} & \small{R} & \small{F1} & $\delta$\small{F1} & $\rho_u$ & $\rho_o$ & $\rho_f$ \\
    \midrule  
    \small{RoBERTa}$^2$ & \small{74.39} & \textbf{\small{76.88}} & \small{75.62} & & \small{7.43} & \small{294} & \small{3.23} \\
    \small{+U,F,O}		& \textbf{\small{75.99}} & \small{76.80} & \textbf{\small{76.39}} & \small{0.8} & \small{4.37} & \small{245} & \small{3.01} \\
    \bottomrule
    \small{Test} & \small{P} & \small{R} & \small{F1} & $\delta$\small{F1} & $\rho_u$ & $\rho_o$ & $\rho_f$ \\
    \midrule  
    \small{RoBERTa}$^2$ & \small{74.79} & \textbf{\small{77.17}} & \small{75.96} & & \small{6.92} & \small{156} & \small{2.67} \\
    \small{+U,F,O}		& \textbf{\small{76.31}} & \small{76.88} & \textbf{\small{76.59}} & \small{0.6} & \small{4.12} & \small{171} & \small{2.41} \\
    \bottomrule
  \end{tabular}
  \caption{Results on low training data ($3$\% of CoNLL-05 and CoNLL-12).
  {RoBERTa}$^2$: Baseline finetuned twice.
  U: Unique core roles. F: Frame core roles. O: Exclusively overlapping roles.
  $\delta$F1: improvement over baseline.
$\rho_f$ is marked NA for the CoNLL-05 test results because ground truth sense is unavailable on the CoNLL-05 shared task page. 
}
  \label{tab:resultlow}
\end{table}

\subsection{Scenario 2: Large Training Data}

\begin{table}[ht!]
  \centering
  \setlength{\tabcolsep}{4pt}
  \renewcommand{\arraystretch}{0.9}
  \begin{tabular}{lccc|c|cc}
  	\multicolumn{7}{c}{\small{CoNLL-05 (100\%, 36k)}} \\
    \toprule
    \small{Dev} & \small{P} & \small{R} & \small{F1} & $\delta$\small{F1} & $\rho_u$ & $\rho_f$ \\
    \midrule  
    \small{RoBERTa}$^2$ & \small{86.74} & \small{87.24} & \small{86.99} & & \small{1.97} & \small{3.23} \\
    \small{+U,F,O }		& \textbf{\small{87.24}} & \textbf{\small{87.26}} & \textbf{\small{87.25}} & \small{0.3} & \small{1.35} & \small{2.99} \\
    \small{Oracle} 	& & & & & \small{0.40} & \small{2.34}  \\
    \bottomrule
    \small{WSJ} & \small{P} & \small{R} & \small{F1} & $\delta$\small{F1} & $\rho_u$ & $\rho_f$ \\
    \midrule  
    \small{RoBERTa}$^2$ & \small{87.75} & \small{87.94} & \small{87.85} & & \small{1.71} & \small{NA} \\
    \small{+U,F,O} 		& \textbf{\small{88.05}} & \textbf{\small{88.00}} & \textbf{\small{88.03}} & \small{0.2} & \small{0.85} & \small{NA} \\
    \small{Oracle} 	& & & & & \small{0.30} & \small{NA}\\
    \bottomrule
    \small{Brown} & \small{P} & \small{R} & \small{F1} & $\delta$\small{F1} & $\rho_u$ & $\rho_f$ \\
    \midrule  
    \small{RoBERTa}$^2$ & \small{79.38} & \small{78.92} & \small{78.64} & & \small{3.36} &  \small{NA} \\
    \small{+U,F,O} 		& \textbf{\small{80.04}} & \textbf{\small{79.56}} & \textbf{\small{79.80}} & \small{1.2} & \small{1.24} & \small{NA} \\
    \small{Oracle} 	& & & & & \small{0.30} & \small{NA}\\
    \bottomrule
  \end{tabular}
  \caption{Results on the \emph{full} CoNLL-05 data.
  Oracle: Errors of oracle.
  $\rho_o$ is in [0,6] across all settings.}
  \label{tab:result05}
\end{table}

Table~\ref{tab:result05}
reports the performance of models trained with our framework using the full training set of the CoNLL-05 dataset
which consists of $35$k sentences with $91$k propositions.
Again, we compare RoBERTa (twice finetuned) with our structure-tuned models.
We see that the constrained models consistently outperform baselines on the dev, WSJ, and Brown sets.
With all three constraints, the constrained model reaches $88$ F1 on the WSJ.
It also generalizes well on new domain by outperforming the baseline by $1.2$ points on
the Brown test set.

As in the low training data experiments, we observe improved precision due to the constraints.  This suggests that even with large
training data, direct label supervision might not be enough for neural models to
pick up the rich output space structure.  Our framework helps neural networks,
even as strong as RoBERTa, to make more correct predictions from differentiable
constraints.

Surprisingly, the development ground truth has a $2.34\%$ error rate on the frame role constraint,
and $0.40\%$ on the unique role constraint.
Similar percentages of unique role errors also appear in WSJ and Brown test sets.
For $\rho_o$, the oracle has no violations on the CoNLL-05 dataset.

The exclusively overlapping constraint (\ie $\rho_o$) is omitted as we found models rarely make such prediction errors.
After adding constraints, the error rate of our model approached the lower bound.
Note that our framework focuses on the learning stage without any specialized
decoding algorithms in the prediction phase except the Viterbi algorithm to
guarantee that there will be no BIO violations.

\paragraph{What about even larger and cleaner data?}
The ideal scenario, of course, is when we have the luxury of massive and clean data to power neural network training.
In Table~\ref{tab:result2012}, we present results on CoNLL-12 which is about $3$ times as large as CoNLL-05.
It consists of $90$k sentences and $253$k propositions.
The dataset is also less noisy with respect to the constraints.
For instance, the oracle development set has no violations for both the unique core and the exclusively overlapping constraints.

We see that, while adding constraints reduced error rates of $\rho_u$ and $\rho_f$,
the improvements on label consistency do not affect F1 much.
As a result, our best constrained model performes on a par with the baseline on the dev set,
and is slightly better than the baseline (by $0.1$) on the test set.
Thus we believe when we have the luxury of data, learning with constraints would become optional. This observation is in line with recent results in \citet{li2019augmenting} and \citet{li2019consistency}.

\paragraph{But is it due to the large data or the strong baseline?}
To investigate whether the seemingly saturated performance is from data or from the model,
we also evaluate our framework on the original BERT~\cite{devlin2019bert} which is relatively less powerful.
We follow the same model setup for experiments and report the performances in Table~\ref{tab:result2012} and Table~\ref{tab:result2012random}.
We see that compared to RoBERTa, BERT obtains similar F1 gains on the test set, suggesting performance ceiling is due to the train size.

\begin{table}[ht!]
  \centering
  \setlength{\tabcolsep}{4pt}
  \renewcommand{\arraystretch}{0.9}
  \begin{tabular}{lccc|c|cc}
  	\multicolumn{7}{c}{\small{CoNLL-12 (100\%, 90k)}} \\
    \toprule
    \small{Dev} & \small{P} & \small{R} & \small{F1} &  $\delta$\small{F1}  & $\rho_u$ & $\rho_f$ \\
    \midrule  
    \small{RoBERTa}$^2$ & \textbf{\small{86.62}} & \textbf{\small{86.91}} & \textbf{\small{86.76}} && \small{0.86} & \small{1.18} \\
    \small{+U,F,O }		& \small{86.60} & \small{86.89} & \small{86.74} & \small{0} & \small{0.59} & \small{1.04} \\
    \small{Oracle} & & & & & \small{0} & \small{0.38}\\
    \bottomrule
    \small{Test} & \small{P} & \small{R} & \small{F1} &  $\delta$\small{F1}  &$\rho_u$ & $\rho_f$ \\
    \midrule  
    \small{RoBERTa}$^2$ & \small{86.28} & \small{86.67} & \small{86.47} && \small{0.91} & \small{0.97} \\
    \small{+U,F,O }		& \textbf{\small{86.40}} & \textbf{\small{86.83}} & \textbf{\small{86.61}} & \small{0.1} & \small{0.50} & \small{0.93} \\
    \small{Oracle} 	& & & & & \small{0} & \small{0.42}\\
    \bottomrule
    \bottomrule
    \small{Dev} & \small{P} & \small{R} & \small{F1} &  $\delta$\small{F1}  & $\rho_u$ & $\rho_f$ \\
    \midrule  
    \small{BERT}$^2$ & \small{85.62} & \small{86.22} & \small{85.92} && \small{1.41} & \small{1.12} \\
    \small{+U,F,O }		& \textbf{\small{85.97}} & \textbf{\small{86.38}} & \textbf{\small{86.18}} & \small{0.3} & \small{0.78} & \small{1.07} \\
    \bottomrule
    \small{Test} & \small{P} & \small{R} & \small{F1} &  $\delta$\small{F1}  &$\rho_u$ & $\rho_f$ \\
    \midrule  
    \small{BERT}$^2$ & \small{85.52} & \small{86.24} & \small{85.88} && \small{1.32} & \small{0.94} \\
    \small{+U,F,O }		& \textbf{\small{85.82}} & \textbf{\small{86.36}} & \textbf{\small{86.09}} & \small{0.2} & \small{0.79} & \small{0.90} \\
    \bottomrule
  \end{tabular}
  \caption{Results on CoNLL-12.
  BERT$^2$: The original BERT finetuned twice.
  $\rho_o$ is around $50$ across all settings.
  With the luxury of large and clean data, constrained learning becomes less effective.}
  \label{tab:result2012}
\end{table}


\section{Ablations \& Analysis}\label{sec:analysis}

In \S\ref{sec:experiments}, we saw that constraints not just improve 
model performance, but also make outputs more structurally consistent. In
this section, we will show the results of an ablation study that adds one
constraint at a time. Then, we will examine the sources of improved F-score by
looking at individual labels, and also the effect of the top-k relaxation for
the constraint $O$. Furthermore, we will examine the robustness of our method against randomness involved during training. We will end this section with a discussion about the ability
of constrained neural models to handle structured outputs. 

\paragraph{Constraint Ablations}\label{sec:howconstrworks}
We present the ablation analysis on our constraints in Table~\ref{tab:ablation}.
We see that as models become more constrained, precision improves.
Furthermore, one class of constraints do not necessarily
reduce the violation rate for the others. Combining all three constraints 
offers a balance between precision, recall, and constraint violation.

One interesting observation that adding the $O$ constraints improve F-scores
even though the $\rho_o$ values were already close to zero.
As noted in \S\ref{sec:overlapconstr}, our constraints apply to the
predicted scores of all labels for a given argument, while the actual decoded
label sequence is just the highest scoring sequence using the Viterbi algorithm.
Seen this way, our regularizers increase the decision
margins on affected labels.  As a result, the model predicts scores that help Viterbi
decoding, and, also generalizes better to new domains \ie, the Brown set.

\begin{table}[h]
  \centering
  \setlength{\tabcolsep}{5pt}
  \renewcommand{\arraystretch}{0.9}
  \begin{tabular}{lccc|cc}
    \multicolumn{6}{c}{\small{CoNLL-05  (100\%, 36k)}} \\
    \toprule
    \small{Dev} & \small{P} & \small{R} & \small{F1} & $\rho_u$ & $\rho_f$ \\
    \midrule  
    \small{RoBERTa}$^2$ & \small{86.74} & \small{87.24} & \small{86.99} & \small{1.97} & \small{3.23} \\
    \small{+U   }   & \small{87.21} & \small{87.32} & \small{87.27} & \small{1.29} & \small{3.23} \\
    \small{+U,F }   & \small{87.19} & \textbf{\small{87.54}} & \textbf{\small{87.37}} & \textbf{\small{1.20}} & \small{3.11} \\
    \small{+U,F,O }   & \textbf{\small{87.24}} & \small{87.26} & \small{87.25} & \small{1.35} & \textbf{\small{2.99}} \\
    \bottomrule
    \small{WSJ} & \small{P} & \small{R} & \small{F1} & $\rho_u$ & $\rho_f$ \\
    \midrule  
    \small{RoBERTa}$^2$ & \small{87.75} & \small{87.94} & \small{87.85} & \small{1.71} & \small{NA} \\
    \small{+U   }   & \small{87.88} & \small{88.01} & \small{87.95} & \small{1.18} & \small{NA} \\
    \small{+U,F }   & \textbf{\small{88.05}} & \textbf{\small{88.09}} & \textbf{\small{88.07}} &  \small{0.89} &\small{NA} \\
    \small{+U,F,O}    & \textbf{\small{88.05}} & \small{88.00} & \small{88.03} & \textbf{\small{0.85}} & \small{NA} \\
    \bottomrule
    \small{Brown} & \small{P} & \small{R} & \small{F1} & $\rho_u$ & $\rho_f$ \\
    \midrule  
    \small{RoBERTa}$^2$ & \small{79.38} & \small{78.92} & \small{78.64} & \small{3.36} &  \small{NA} \\
    \small{+U   }   & \small{79.36} & \small{79.15} & \small{79.25} & \small{1.74} & \small{NA} \\
    \small{+U,F }   & \small{79.60} & \small{79.24} & \small{79.42} & \textbf{\small{1.00}} & \small{NA} \\
    \small{+U,F,O}    & \textbf{\small{80.04}} & \textbf{\small{79.56}} & \textbf{\small{79.80}} & \small{1.24} & \small{NA} \\
    \bottomrule
  \end{tabular}
  \caption{Ablation tests on CoNLL-05.}
  \label{tab:ablation}
\end{table}

\paragraph{Sources of Improvement}\label{sec:sourceofimprovement}
Table~\ref{tab:labelf1} shows label-wise F1 scores for each argument.
Under low training data conditions, our constrained models gained improvements primarily from
the frequent labels, \eg, \lbl{A0}-\lbl{A2}.
On CoNLL-05 dataset, we found the location modifier (\lbl{AM-LOC}) posed challenges to our constrained models
which significantly performed worse than the baseline.
Another challenge is the negation modifier (\lbl{AM-NEG}), where our models underperformed on both datasets,
particularly with small training data.
When using the CoNLL-12 training set, our models performed on par with the baseline even on frequent labels,
confirming that the performance of soft-structured learning is nearly saturated on the larger, cleaner dataset.

\begin{table*}[ht!]
  \centering
  \setlength{\tabcolsep}{4pt}
  \renewcommand{\arraystretch}{0.8}
  \begin{tabular}{l|cc|cc|cc|cc}
    & \multicolumn{2}{c|}{\small{CoNLL-05 3\%}} & \multicolumn{2}{c|}{\small{CoNLL-05 100\%}} & \multicolumn{2}{c}{\small{CoNLL-12 3\%}} & \multicolumn{2}{c}{\small{CoNLL-12 100\%}} \\
    & \small{RoBERTa}$^2$ & \small{+U,F,O} & \small{RoBERTa}$^2$ & \small{+U,F,O} & \small{RoBERTa}$^2$ & \small{+U,F,O} & \small{RoBERTa}$^2$ & \small{+U,F,O} \\
    \toprule
    \small{\lbl{A0}    } & \small{81.28} & \small{82.11} & \small{93.43} & \small{93.52} & \small{84.99} & \small{85.73} & \small{92.78} & \small{92.81} \\
    \small{\lbl{A1}    } & \small{72.12} & \small{73.59} & \small{89.23} & \small{89.80} & \small{78.36} & \small{79.67} & \small{89.88} & \small{89.75} \\
    \small{\lbl{A2}    } & \small{46.50} & \small{47.52} & \small{79.53} & \small{79.73} & \small{68.24} & \small{69.20} & \small{84.93} & \small{84.90} \\
    \small{\lbl{A3}    } & \small{39.58} & \small{42.11} & \small{81.45} & \small{81.86} & \small{33.26} & \small{34.47} & \small{72.96} & \small{73.24} \\
    \small{\lbl{A4}    } & \small{51.61} & \small{51.56} & \small{74.60} & \small{75.59} & \small{56.29} & \small{58.38} & \small{80.80} & \small{80.33} \\
    \small{\lbl{AM-ADV}} & \small{44.07} & \small{47.56} & \small{66.67} & \small{66.91} & \small{55.26} & \small{54.93} & \small{66.37} & \small{66.92} \\
    \small{\lbl{AM-DIR}} & \small{16.39} & \small{18.92} & \small{55.26} & \small{55.56} & \small{36.51} & \small{35.81} & \small{64.92} & \small{64.95} \\
    \small{\lbl{AM-DIS}} & \small{71.07} & \small{70.84} & \small{80.20} & \small{80.50} & \small{76.35} & \small{76.40} & \small{82.86} & \small{82.71} \\
    \small{\lbl{AM-LOC}} & \small{53.08} & \small{51.60} & \small{69.02} & \small{66.50} & \small{59.74} & \small{59.94} & \small{72.74} & \small{73.21} \\
    \small{\lbl{AM-MNR}} & \small{44.30} & \small{44.18} & \small{68.63} & \small{69.87} & \small{56.14} & \small{55.67} & \small{70.89} & \small{71.13} \\
    \small{\lbl{AM-MOD}} & \small{91.88} & \small{91.60} & \small{98.27} & \small{98.60} & \small{95.50} & \small{95.76} & \small{97.88} & \small{98.04} \\
    \small{\lbl{AM-NEG}} & \small{91.18} & \small{88.35} & \small{94.06} & \small{93.60} & \small{93.29} & \small{93.05} & \small{95.93} & \small{95.83} \\
    \small{\lbl{AM-TMP}} & \small{74.05} & \small{74.13} & \small{88.24} & \small{88.08} & \small{79.00} & \small{78.78} & \small{87.58} & \small{87.56} \\
    \midrule  
    \small{Overall}  & \small{70.48} & \small{71.55} & \small{87.33} & \small{87.61} & \small{76.66} & \small{77.45} & \small{87.60} & \small{87.58} \\
    \bottomrule
  \end{tabular}
  \caption{Label-wise F1 scores for the CoNLL-05 and CoNLL-12 development sets.}
  \label{tab:labelf1}
\end{table*}

\paragraph{Impact of Top-$k$ Beam Size} \label{sec:topk}
As noted in \S\ref{sec:overlapconstr}, we used the top-$k$ strategy to implement the constraint $O$.
As a result, there is a certain chance for predicted label sequences to have non-exclusive overlap without our regularizer
penalizing them.
What we want instead is a good balance between coverage and runtime cost.
To this end, we analyze the CoNLL-12 development set using the baseline trained on $3\%$ of CoNLL-12 data.
Specifically, we count the examples which have such overlap but the regularization loss is $\leq 0.001$.
In Table~\ref{tab:topk}, we see that $k=4$ yields good coverage.

\begin{table}[h]
  \centering
  \renewcommand{\arraystretch}{0.9}
  \begin{tabular}{lccccc}
    \toprule
    {k} & {1} & {2} & {4} & {6} \\
    {\# Ex.} & {10} & {8} & {3} & {2} \\
    \bottomrule
  \end{tabular}
  \caption{Impact of $k$ for the top-$k$ strategy, showing the number of missed examples for different $k$.
  We set $k=4$ across all experiments.}
  \label{tab:topk}
\end{table}

\paragraph{Robustness to random initialization}
We observed that model performance with structured tuning is generally robust to random initialization. As an illustration, we show the performance of models trained on the full CoNLL-12 dataset with different random initializations in Table~\ref{tab:result2012random}. 

\begin{table}[ht!]
  \centering
  \setlength{\tabcolsep}{6pt}
  \renewcommand{\arraystretch}{0.9}
  \begin{tabular}{lccc|c}
    \multicolumn{5}{c}{\small{CoNLL-12 (100\%, 90k)}} \\
    \toprule
    \small{Test F1} & \small{Seed1} & \small{Seed2} & \small{Seed3} & \small{avg }$\delta$\small{F1} \\
    \midrule  
    \small{BERT}$^2$ & \small{85.88} & \small{85.91} & \small{86.13} & \\
    \small{+U,F,O }   & \textbf{\small{86.09}} & \textbf{\small{86.07}} & \textbf{\small{86.19}} &  \small{0.1} \\
    \bottomrule
    \bottomrule
    \small{Test F1} & \small{Seed1} & \small{Seed2} & \small{Seed3} & \small{avg }$\delta$\small{F1}  \\
    \midrule  
    \small{RoBERTa}$^2$ & \small{86.47} & \small{86.33} & \small{86.45} & \\
    \small{+U,F,O }   & \textbf{\small{86.61}} & \textbf{\small{}86.48} & \textbf{\small{86.57}} & \small{0.1} \\
    \bottomrule
  \end{tabular}
  \caption{F1 scores models trained on the CoNLL-12 data with different random seeds. The randomness affects the initialization of the classification layers and the batch ordering during training.}
  \label{tab:result2012random}
\end{table}

\paragraph{Can Constrained Networks Handle Structured Prediction?}
Larger, cleaner data may presumably be better for training constrained neural
models.  But it is not that simple.  We will approach the above question by
looking at how good the transformer models are at dealing with two classes of constraints,
namely: 1) structural constraints that rely \emph{only}  on available decisions
(constraint $U$), 2) constraints involving external knowledge (constraint $F$).

For the former, we expected neural models to perform very well since the
constraint $U$ represents a simple local pattern.  From
Tables~\ref{tab:result05} and~\ref{tab:result2012}, we see that the constrained
models indeed reduced violations $\rho_u$ substantially.  However, when the
training data is limited, \ie, comparing CoNLL-05 $3\%$ and $100\%$, the
constrained models, while reducing the number of errors, still make many invalid
predictions.  We conjecture this is because networks learn
with constraints mostly by memorization.  Thus the ability to generalize learned
patterns on unseen examples relies on training size.

The constraint $F$ requires external knowledge from the
PropBank frames.  We see that even with large training data, constrained models
were only able to reduce error rate $\rho_f$ by a small margin.  In our
development experiments, having larger $\lambda_F$ tends to strongly sacrifice
argument F1, yet still does not to improve development error rate
substantially.  Without additional training signal in the form of such background
knowledge, constrained inference becomes a necessity, even with strong neural
network models.


\section{Discussion \& Conclusion}\label{sec:related}

\paragraph{Semantic Role Labeling \& Constraints}

The SRL task is inherently knowledge rich; the outputs are defined in terms of
an external ontology of frames. The work presented here can be generalized to
several different flavors of the task, and indeed, constraints could be used to
model the interplay between them. For example, we could revisit the analysis of
\citet{yi-etal-2007-semantic}, who showed that the PropBank \lbl{A2} label takes
on multiple meanings, but by mapping them to VerbNet, they can be
disambiguated. Such mappings naturally define constraints that link semantic
ontologies.

Constraints have long been a cornerstone in the SRL models.  Several early
linear models for SRL
\cite[e.g.][]{punyakanok2004semantic,punyakanok2008importance,surdeanu2007combination}
modeled inference for PropBank SRL using integer linear programming.
\citet{riedel2008collective} used Markov Logic Networks to learn and predict
semantic roles with declarative constraints. The work of
\cite{tackstrom2015efficient} showed that certain SRL constraints admit
efficient decoding, leading to a neural model that used this
framework~\cite{fitzgerald2015semantic}.  Learning with constraints has also
been widely adopted in semi-supervised SRL~\cite[e.g.,][]{furstenau2012semi}.

With the increasing influence of neural networks in NLP, however, the role of
declarative constraints seem to have decreased in favor of fully end-to-end
training~\cite[e.g.,][and others]{he2017deep,strubell2018linguistically}. In
this paper, we show that even in the world of neural networks with contextual
embeddings, there is still room for systematically introducing knowledge in the
form of constraints, without sacrificing the benefits of end-to-end learning.

\paragraph{Structured Losses}
\citet{chang2012structured} and \citet{ganchev2010posterior} developed models
for structured learning with declarative constraints.  Our work is in the same
spirit of training models that attempts to maintain output consistency.

There are some recent works on the design of models and loss functions by relaxing
Boolean formulas. \citet{kimmig2012short} used the \L{}ukasiewicz t-norm for
probabilistic soft logic.  \citet{li2019augmenting} augment the neural network
architecture itself using such soft logic. \citet{pmlr-v80-xu18h} present a general framework for
loss design that does not rely on soft logic.  Introducing extra regularization terms
to a downstream task have been shown to be beneficial in terms of both output
structure consistency and prediction
accuracy~\cite[e.g.,][]{minervini2018adversarially,hsu2018unified,mehta2018towards,du2019consistent,li2019consistency}.

\paragraph{Final words} In this work, we have presented a framework that seeks
to predict structurally consistent outputs without extensive model
redesign, or any expensive decoding at prediction time. Our experiments on the
semantic role labeling task show that such an approach can be especially
helpful in scenarios where we do not have the luxury of massive annotated
datasets.



\section*{Acknowledgements}\label{sec:acknowledge}

We thank members of the NLP group at the University of Utah for their valuable
insights and suggestions; and reviewers for pointers to related works,
corrections, and helpful comments. We also acknowledge the support of NSF
Cyberlearning-1822877, SaTC-1801446, U.S. DARPA KAIROS Program No. FA8750-19-2-1004, 
DARPA Communicating with Computers DARPA 15-18-CwC-FP-032, HDTRA1-16-1-0002, 
and gifts from Google and NVIDIA.

The views and conclusions contained herein are those of the authors and 
should not be interpreted as necessarily representing the official policies, 
either expressed or implied, of DARPA or the U.S. Government. The U.S. 
Government is authorized to reproduce and distribute reprints for governmental 
purposes notwithstanding any copyright annotation therein.


\bibliographystyle{acl_natbib}
\bibliography{cited}

\appendix
\section{Appendices}
\label{sec:appendix}

\subsection{Hyperparameters}
We show the hyperparameters of $\lambda$`s in Table~\ref{tab:lambdas}.
We conducted grid search on the combinations of $\lambda$`s for each setting and the best one on development set is selected for reporting.

\begin{table}[ht!]
  \centering
  \renewcommand{\arraystretch}{0.9}
  \begin{tabular}{lccc}
    Model & \small{$\lambda_U$} & \small{$\lambda_O$} & \small{$\lambda_F$} \\
    \midrule
    \small{RoBERTa CoNLL-05 (3\%)}  \\
    \small{+U,F,O} & 2 & 0.5 & 0.5 \\
    \midrule
    \small{RoBERTa CoNLL-2012 (3\%)}  \\
    \small{+U,F,O} & 1 & 2 & 1 \\
    \midrule
    \small{RoBERTa CoNLL-05 (100\%)}  \\
    \small{+U} & 1 \\
    \small{+U,F} & 1 & 0.5 \\
    \small{+U,F,O} & 1 & 0.5 & 0.1 \\
    \midrule
    \small{RoBERTa CoNLL-2012 (100\%)}  \\
    \small{+U,F,O} & 1 & 1 & 0.1 \\
    \midrule
    \small{BERT CoNLL-2012 (100\%)}  \\
    \small{+U,F,O} & 0.5 & 1 & 0.1 \\
    \bottomrule
  \end{tabular}
  \caption{Values of hyperparameter $\lambda$`s.}
  \label{tab:lambdas}
\end{table}

\end{document}